\documentclass[journal]{IEEEtran}
\usepackage[T1]{fontenc}
\usepackage{cite}
\usepackage{url}
\usepackage{amsmath,amssymb,amsfonts}
\usepackage{graphicx}
\usepackage{textcomp}
\usepackage{xcolor}
\usepackage{hyperref}
\usepackage{bm}
\usepackage{amsthm}
\usepackage{enumerate}
\usepackage{booktabs}
\usepackage{diagbox}
\usepackage{supertabular}
\usepackage{subfigure}
\usepackage{caption}
\usepackage{algorithmicx}
\usepackage[linesnumbered,ruled,vlined]{algorithm2e}

\begin{document} 

\title{From Generative AI to Generative Internet of Things: Fundamentals, Framework, and Outlooks}

\author{Jinbo Wen, Jiangtian Nie, Jiawen Kang, Dusit Niyato, \textit{IEEE Fellow}, Hongyang Du,\\ Yang Zhang, Mohsen Guizani, \textit{IEEE Fellow}
\thanks{
J. Wen and Y. Zhang are with the College of Computer Science and Technology, Nanjing University of Aeronautics and Astronautics, China (e-mail: jinbo1608@163.com; yangzhang@nuaa.edu.cn). J. Nie, D. Niyato, and H. Du are with the School of Computer Science and Engineering, Nanyang Technological University, Singapore (e-mail: jnie001@e.ntu.edu.sg; dniyato@ntu.edu.sg; hongyang001@e.ntu.edu.sg). J. Kang is with the School of Automation, Guangdong University of Technology, China (e-mail: kavinkang@gdut.edu.cn). M. Guizani is with the Mohamed bin Zayed University of Artificial Intelligence (MBZUAI), UAE (e-mail: mguizani@ieee.org).
\textit{*Corresponding author: Yang Zhang.}
}
}

\maketitle

\begin{abstract}

 
Generative Artificial Intelligence (GAI) possesses the capabilities of generating realistic data and facilitating advanced decision-making. By integrating GAI into modern Internet of Things (IoT), Generative Internet of Things (GIoT) is emerging and holds immense potential to revolutionize various aspects of society, enabling more efficient and intelligent IoT applications, such as smart surveillance and voice assistants. In this article, we present the concept of GIoT and conduct an exploration of its potential prospects. Specifically, we first overview four GAI techniques and investigate promising GIoT applications. Then, we elaborate on the main challenges in enabling GIoT and propose a general GAI-based secure incentive mechanism framework to address them, in which we adopt Generative Diffusion Models (GDMs) for incentive mechanism designs and apply blockchain technologies for secure GIoT management. Moreover, we conduct a case study on modern Internet of Vehicle traffic monitoring, which utilizes GDMs to generate effective contracts for incentivizing users to contribute sensing data with high quality. Numerical results demonstrate the superiority of the proposed scheme. Finally, we suggest several open directions worth investigating for the future popularity of GIoT.
\end{abstract}


\begin{figure*}[t]
\centerline{\includegraphics[width=0.95\textwidth]{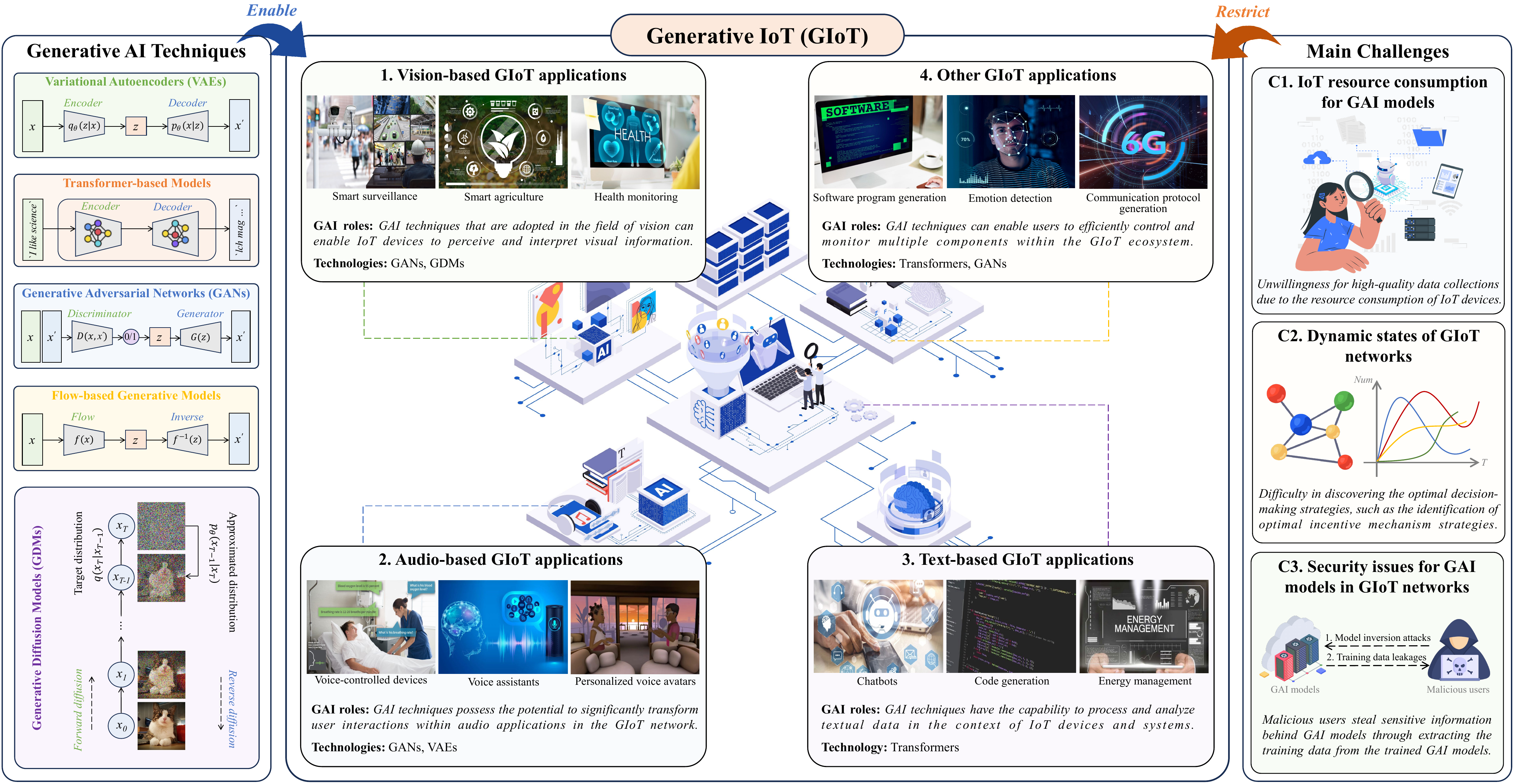}}
\captionsetup{font=footnotesize}
\caption{The schematic of generative IoT networks. We summarize four potential generative IoT applications, encompassing vision-based, audio-based, and text-based applications. Additionally, we discuss several GAI techniques that enable generative IoT applications and identify the main challenges that restrict the development and widespread adoption of such applications.}
\label{system}
\end{figure*}

\section{Introduction}
The advent of Generative Artificial Intelligence (GAI) represents a significant milestone in the field of AI\cite{du2023beyond}. In contrast to traditional AI models that primarily classify or analyze existing data, GAI possesses the incredible ability to generate novel content such as digital films, audio, photos, or codes, thus exerting profound impacts across various domains\cite{zhang2023complete}. For instance, in the healthcare domain, GAI can assist physicians in diagnosing conditions based on medical records and images, and generate tailored treatment plans for patients. In the tourism and hospitality domain, GAI can generate hyper-personalized content for tourists, fostering changes in tourism strategies such as destination planning and hotel booking. Simultaneously, the potential of GAI for network optimization has been explored\cite{du2023beyond}, contributing to the optimization of network management and performance, thereby enhancing the efficiency of decision-making in complex networks\cite{du2023beyond}.

Recent advances in cutting-edge technologies, such as sixth-generation (6G) wireless communications, Artificial Intelligence (AI), and edge computing, are bringing modern Internet of Things (IoT) technologies to maturity\cite{9509294}. Modern IoT is considered an intelligent and autonomous ecosystem that revolutionizes device connectivity, data analytics, and intelligent decision-making. With its capabilities of ultra-low latency communications, seamless connectivity, and ubiquitous computing\cite{9509294}, modern IoT has the potential to enable novel and advanced applications across various industries and domains, including intelligent healthcare monitoring and smart homes. For instance, in smart homes, IoT devices can automate and control various aspects of home living, such as smart lighting systems and temperature control, providing enhanced convenience and comfort for users.

Given the remarkable capabilities of GAI in generating realistic data and facilitating advanced decision-making processes\cite{zhang2023generative}, 
we envision that GAI-empowered modern IoT will become more creative and proactive as such the term \textit{Generative IoT (GIoT)} emerges. 
By leveraging the advanced data generation and decision-making capabilities of GAI, GIoT has the capacity to drive the progression of IoT-enabled environments.
For instance, by processing historical sensing data and real-time sensor readings, GAI can forecast future events, predict system failures, and generate effective resource management to improve overall system performance\cite{du2023beyond}. Although GIoT holds significant potential to revolutionize various domains, there exist several critical challenges that need to be addressed when integrating GAI with modern IoT to enable GIoT, including the impact of IoT resource consumption on the performance of GAI model fine-tuning, the dynamic nature of GIoT networks complicating the identification of optimal decision strategies, and security concerns for GAI models in GIoT networks. \textit{To the best of our knowledge, this is the first work that presents the concept of GIoT and systemically provides foresight research on the integration of GAI with modern IoT for enabling GIoT}. The contributions of this article can be summarized as follows:
\begin{itemize}
    \item We first provide a comprehensive overview of GAI techniques that have been widely adopted in the field of computer vision. Then, we systematically discuss the potential GIoT applications and the main challenges of synergy between GAI and modern IoT to enable GIoT.
    \item We present a general incentive mechanism framework to address the main challenges in enabling GIoT. We utilize blockchain technologies to manage and secure GIoT and adopt Generative Diffusion Models (GDMs) to derive optimal incentive mechanism design.
    \item We conduct a case study on modern Internet of Vehicles (IoV) traffic monitoring, in which we develop a GDM-based contract theory model for incentivizing users to contribute high-quality sensing traffic data. Numerical results demonstrate that the edge server utility of the proposed scheme is almost $52\%$ higher than that of the Deep Reinforcement Learning (DRL)-based scheme.
\end{itemize}

\section{Generative Internet of Things}
In this section, we introduce several widely adopted GAI techniques, especially in the field of computer vision, involving their basic architectures and applications for modern IoT. Then, we systematically explore the potential GIoT applications. Finally, we discuss the main challenges posed by enabling GIoT, as shown in Fig. \ref{system}.

\renewcommand{\arraystretch}{1.2}
\begin{table*}[t]
\centering
\caption{Advantage Comparisons of Generative AI and Traditional AI for Significant IoT Applications.}
\begin{tabular}{m{3.0cm}|m{7.3cm}|m{5.8cm}}
\toprule[1pt]
\hline
\multicolumn{1}{c|}{\textbf{IoT Applications}}& \multicolumn{2}{c}{\textbf{Advantages for the Applications}} \\ \cline{2-3}
& \multicolumn{1}{c|}{\textbf{Generative AI}} & \multicolumn{1}{c}{\textbf{Traditional AI}} \\ \hline
Vision-based applications & \begin{itemize}
    \item \textit{Advanced image recognition:} GAI can attain higher accuracy in image recognition tasks by learning complex visual patterns from extensive datasets\cite{ferrag2023generative}.
    \item \textit{Enhanced object detection and tracking:} GAI can accurately detect and track objects without explicit labeling in real-time, enabling more robust surveillance systems\cite{zhang2023complete}.
    \item \textit{Intelligent semantic understanding}: GAI can go beyond basic image processing and understand the semantic context of images, enabling more advanced applications like scene understanding and image captioning\cite{xia2023generative}.
\end{itemize} & \begin{itemize}
    \item \textit{Efficient vision processing:} Traditional AI can be computationally less demanding compared to GAI, making them more efficient for specific vision-based tasks, particularly in large-scale image processing tasks.
    \item \textit{Simpler vision architectures:} Traditional AI, such as classical computer vision techniques, often utilize simpler architectures, e.g., convolutional neural networks and support vector machines, which are easier to implement.
\end{itemize}
\\ \hline
Audio-based applications & 
\begin{itemize}
    \item \textit{Accurate speech recognition:} GAI can achieve higher accuracy in speech recognition tasks even in noisy acoustic environments, enabling voice-controlled IoT devices\cite{cao2023comprehensive}.
    \item \textit{Enhanced voice understanding:} GAI can interpret spoken language by enhancing context and semantics, enabling intelligent voice assistants\cite{ferrag2023generative}.
    \item \textit{Smart sound anomaly detection:} GAI can detect anomalous audio patterns, enabling IoT applications like acoustic surveillance and predictive maintenance\cite{zhang2023complete}.
\end{itemize}
& \begin{itemize}
    \item \textit{Robust noise handling:} Traditional AI can effectively handle noisy audio signals since they are often designed with specific signal processing techniques, such as filtering and noise cancellation.
    \item \textit{Tailored audio intelligence:} Traditional AI can be tailored to specific audio-based tasks, such as speaker identification and audio classification, demonstrating high performance.
\end{itemize} \\ \hline
Text-based applications & \begin{itemize}
    \item \textit{Nuanced text analysis:} GAI can process and analyze text data with more nuanced language understanding, enabling sentiment analysis and language translation\cite{cao2023comprehensive}.
    \item \textit{Intelligent contextual understanding:} GAI can capture profound meaning from context, enabling more intelligent chatbots and personalized content delivery\cite{zhang2023complete}.
    \item \textit{Precise code assistance:} GAI can generate programming codes from natural language description and provide coding assistance for users\cite{cao2023comprehensive}.
\end{itemize} & \begin{itemize}
    \item \textit{Efficient rule-based processing:} Traditional AI, such as rule-based systems, can be effective for text-based applications that require explicit and rule-driven decision-making.
    \item \textit{Transparent decision insight:} Traditional AI can enable users to understand the decision-making process and analyze the logic behind the output results.
\end{itemize} \\ \hline
\bottomrule[1pt]
\end{tabular}
\label{GAI_AI}
\end{table*}
\renewcommand{\arraystretch}{1}

\subsection{Generative AI Technologies}
As a powerful branch of AI, GAI focuses on creating new content in various modalities, such as videos, images, text, and audio\cite {zhang2023generative}. GAI can leverage pre-trained models to generate new content by fine-tuning the model parameters based on user-provided input, i.e., prompts. Moreover, GAI can utilize learning algorithms to automate content generation from existing data. Motivated by recent studies\cite{cao2023comprehensive,du2023beyond}, we introduce four widely adopted model-based GAI techniques\cite{cao2023comprehensive} and explore their potential for IoT applications. 
\begin{itemize}
    \item \textbf{Variational Autoencoders (VAEs):} VAEs consist of the encoder and decoder networks\cite{cao2023comprehensive}. The encoder network compresses the input data to a latent representation. Then, the decoder network learns to reconstruct synthetic data that closely aligns with the original distribution\cite{liu2023deep}. 
    Due to their ability to effectively represent data in a probabilistic latent space, VAEs can be applied to various IoT applications, such as energy optimization and equipment maintenance. For instance, by training on real-time sensor readings, VAEs can capture complex data distributions and generate more robust predictions on future equipment conditions than traditional AI methods\cite{cook2019anomaly}.
    
    \item \textbf{Generative Adversarial Networks (GANs):} GANs have been applied widely in IoT data synthesis, consisting of generator and discriminator networks. The generator network aims to generate new data by learning real data distribution, while the discriminator network aims to distinguish synthetic data from real data\cite{liu2023deep}. The two networks are trained together in interactive and competitive manners, resulting in continuous enhancement of synthesis performance. 
    With good performance in generating realistic samples\cite{liu2023deep}, GANs can be utilized not only for data augmentation but also for IoT anomaly detection\cite{cook2019anomaly}. Notably, unlike traditional AI methods that typically require retraining on labeled data to adapt to changes, GANs can learn the underlying data distribution in an unsupervised manner, enabling the adaptation to evolving anomalies without explicit labeling.
    
    
    \item \textbf{Flow-based Generative Models (FGMs):} FGMs can transform input data distributions from simple to complex through a series of differentiable and invertible transformations that are implemented as neural networks\cite{cao2023comprehensive}. 
    Unlike VAEs and GANs, FGMs possess the distinctive capability to learn explicitly the data distribution and directly compute the probability density function during generation \cite{du2023beyond}. Therefore, FGMs can circumvent resource-intensive computation and directly model complex probability distributions, which can be effectively applied in IoT domains such as traffic flow optimization\cite{xu2023unleashing} and anomaly detection in network traffic.
    

    \item \textbf{Generative Diffusion Models (GDMs):} 
    With the state-of-the-art performance of image synthesis, GDMs are emerging generative models\cite{cao2023comprehensive}, 
    consisting of forward diffusion and denoising processes inspired by non-equilibrium thermodynamics theory\cite{liu2023deep}. 
    Because of their recent advancements in training and sampling efficiency, GDMs have been used not only for image generation but also for IoT network optimizations\cite{du2023beyond}. Specifically, GDMs exhibit the ability to capture complex and high-dimensional structures, effectively addressing network optimization problems and decision-making processes\cite{du2023beyond}, while traditional AI methods often converge slowly and stuck in locally optimal solutions.

\end{itemize}

\subsection{Generative IoT Applications}
Unlike traditional AI, GAI with its capability of generating realistic and context-aware data has the potential to revolutionize various industries, such as healthcare, manufacturing, transportation, and smart cities. Note that the detailed advantage comparison of GAI and traditional AI for significant IoT applications is listed in Table \ref{GAI_AI}. With the incorporation of GAI into modern IoT systems, a new paradigm called GIoT is emerging. GIoT holds significant potential for transformative applications across various domains. By capitalizing on advanced IoT and GAI technologies, GIoT can enable intelligent systems, optimize resource utilization, enhance decision-making processes, and improve overall efficiency and sustainability in diverse sectors.  

\textbf{\emph{1) Vision-based GIoT applications:}} Vision-based GIoT applications leverage the power of GAI technologies, particularly GAI techniques for computer vision like GDMs and GANs, to enable IoT devices to perceive and interpret visual information. Vision-based applications can be applied in various types of GIoT contexts, from real-time monitoring to remote diagnostic and maintenance\cite{ferrag2023generative}, such as smart surveillance, smart agriculture, and health monitoring. For example, in smart surveillance systems, unlike the limited generalization capability of traditional AI, GAI can leverage data collected by IoT devices equipped with cameras and smart sensors to track objects and detect variable suspicious activities in various domains.
Besides, in smart agriculture, GAI-empowered IoT devices can revolutionize farming practices. Specifically, since traditional AI lacks adaptability to changing environmental conditions, GAI models can be adopted to predict the growth of crops based on the crop data captured by cameras mounted on drones or ground-based sensors, and the predictions can be shown in the form of images or videos\cite{ferrag2023generative}, which facilitates data-driven decision-making for farmers.

\textbf{\emph{2) Audio-based GIoT applications:}}
GAI technologies can advance our interactions with audio applications in GIoT networks. One of the most notable examples of audio applications based on GAI is voice assistant. Voice assistants, such as Amazon Alexa\footnote{\url{https://www.aboutamazon.com/news/devices/amazon-alexa-generative-ai}} and Apple Siri\footnote{\url{https://appleinsider.com/articles/23/09/06/}} that utilize audio-based GAI techniques, can understand and respond to voice commands like adjusting room thermostats and turning on and off lamps. Another important application of GAI in the audio domain is creating personalized audio systems for avatars that are highly accurate digital replicas of users, such as Resemble AI\footnote{\url{https://www.resemble.ai/}}. By analyzing the preferences and characteristics of users, GAI models can generate tailored audio systems for their avatars, seamlessly immersing users, especially in metaverses. 

\textbf{\emph{3) Text-based GIoT applications:}}
Text-based GIoT applications involve leveraging GAI models like ChatGPT\footnote{\url{https://chat.openai.com/}} to process and analyze text data in the context of IoT devices and systems. Specifically, GAI-empowered chatbots can process textual data generated by IoT devices and provide real-time insights. These conversational agents employ natural language processing algorithms to interpret natural language input, generate responses to users, and perform specific tasks like controlling IoT devices. Additionally, text-based GIoT applications also involve automated code generation. Specifically, by analyzing high-level specifications of desired functionality, GAI models, such as Codex\footnote{\url{https://openai.com/blog/openai-codex}} as a general-purpose programming model created by OpenAI, can automatically generate corresponding codes for specific IoT applications \cite{ferrag2023generative}.


\textbf{\emph{4) Other GIoT applications:}}
There are also other novel GIoT applications in different modalities based on GAI technologies. One potential application is the automated generation of software programs, which can be downloaded to various IoT devices, enabling users to efficiently control and monitor multiple components within the GIoT ecosystem. Another potential application is the generation of secure communication protocols\cite{ferrag2023generative}, such as the 6G wireless communication protocol. Since wireless communications between IoT devices are vulnerable to being compromised by malicious attackers, GAI techniques can be utilized to develop robust communication protocols that encrypt the data transmitted between devices, making attackers more difficult to access critical data in transit.

\begin{figure*}[t]
\centerline{\includegraphics[width=0.95\textwidth]{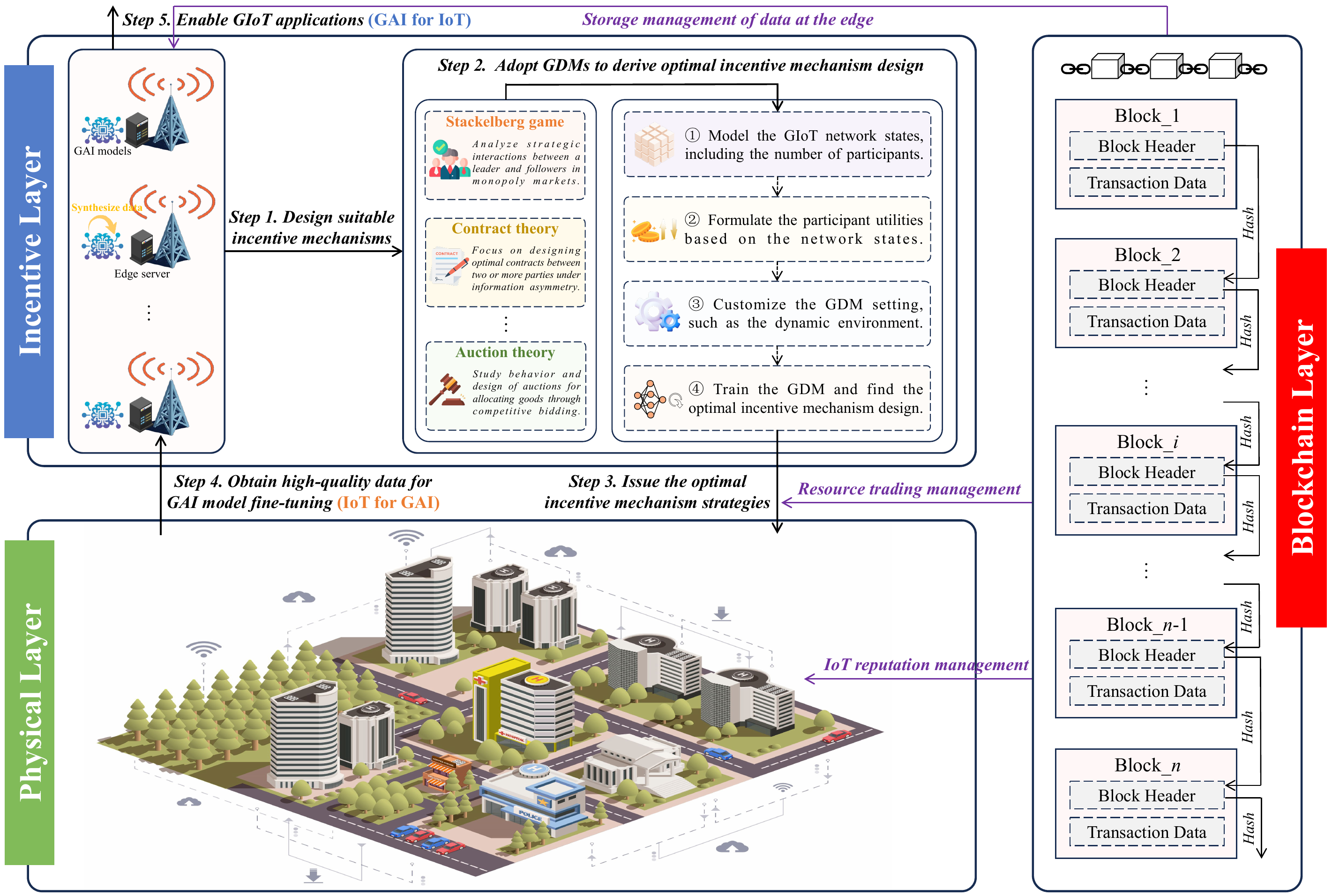}}
\captionsetup{font=footnotesize}
\caption{Generative AI-based secure incentive mechanism framework for generative IoT. The proposed framework consists of three layers, i.e., a physical layer, an incentive layer, and a blockchain layer, where the incentive layer is used to motivate users with IoT devices to provide high-quality data for generative AI model fine-tuning, and the blockchain layer is used to securely manage generative IoT networks.}
\label{framework}
\end{figure*}

\subsection{Main Challenges in Integrating GAI with Modern IoT}
Although GAI technologies hold great potential for transforming the modern IoT ecosystem, the convergence of GAI technologies with modern IoT still suffers from the following challenges, which should be resolved for the future popularization and development of GIoT.

\emph{\textbf{C1) IoT resource consumption for GAI models:}}
In GIoT networks, GAI models require a modest quantity of extra data to perform model fine-tuning and direct inference at the edge, minimizing service latency and enhancing user experiences\cite{xu2023unleashing}. The data for model fine-tuning can be generated in the cloud or collected by IoT devices and subsequently uploaded by mobile users\cite{xu2023unleashing}. However, if the dataset contains deviations and inaccuracies in information, pre-trained GAI models cannot be accurately fine-tuned to specific tasks or domains, leading to inaccurate and biased inferences. Therefore, the dataset needs to be high-quality to avoid incorrect learning patterns in the GAI model fine-tuning\cite{wen2023freshness}. 
Since data collection and transmission lead to high costs, users may not be reluctant to contribute high-quality data to the edge due to the resource constraints of IoT devices, affecting the performance of GAI model fine-tuning.

\emph{\textbf{C2) Dynamic states of GIoT networks:}}
Due to the scale and complexity of interconnected devices as well as the dynamic and real-time nature of the network, GIoT can be considered as a heterogeneous and large-scale system\cite{luong2016data}. Consequently, intricate decision-making processes arise, such as the optimal allocation of limited IoT network resources and the identification of optimal incentive mechanism strategies. Generally, the optimal decision-making strategies are determined by employing the traditional optimization principle and tools\cite{du2023beyond}. However, these approaches often rely on accurate and comprehensive network information, which are not feasible in complex GIoT network scenarios. Additionally, while DRL has shown promise in various network optimization and decision-making tasks, the dynamics of IoT networks can significantly impact the state and action spaces of DRL models. This necessitates the complete retraining of DRL models\cite{liu2023deep}, which may inefficiently discover the optimal decision-making strategies in dynamic GIoT network scenarios.


\emph{\textbf{C3) Security issues for GAI models in GIoT networks:}}
The heterogeneity of GIoT networks, exemplified by the ability of IoT devices to dynamically join or leave the networks as required\cite{luong2016data}, poses a significant difficulty in secure management for GIoT networks. Ensuring the quality and diversity of collected data by IoT devices is one of the key challenges. Specifically, malicious users equipped with IoT devices would deliberately upload low-quality data to the edge to obtain more benefits. Additionally, malicious users can issue model inversion attacks to steal sensitive information behind GAI models by extracting the training data from trained models\cite{ferrag2023generative}. For example, based on the text generated by ChatGPT, malicious users can deduce private information from either the fine-tuning data or the data employed to pre-train the foundation model of ChatGPT, which may lead to serious security threats to other normal users. 

Motivated by the above analysis, it is necessary to develop a reliable and secure incentive mechanism framework, thereby enabling more intelligent and autonomous GIoT ecosystems. The proposed framework is discussed in Section \ref{pro_fra}.

\section{Generative AI-based Incentive Mechanism Framework for Generative IoT}\label{pro_fra}
In this section, we introduce several representative techniques for designing incentive mechanisms for GIoT. To address the aforementioned challenges, we propose a general GAI-based secure incentive mechanism framework.

\subsection{Incentive Mechanism Design for Generative IoT}
In IoT network optimizations, incentive mechanisms play a crucial role in incentivizing network users to actively contribute their resources, share data, or collaborate, thereby improving the performance and reliability of the network\cite{du2023beyond}. In the following part, we discuss several representative techniques for developing incentive mechanisms, i.e., Stackelberg game\cite{zhong2023blockchain}, contract theory\cite{wen2023freshness}, and auction theory\cite{zhang2020auction}, which have been widely adopted in IoT network optimizations\cite{du2023beyond}.
\subsubsection{Stackelberg game}
As a non-cooperative game theory, the Stackelberg game focuses on analyzing strategic interactions between a leader and followers, especially in IoT networks, where the leader first determines resource prices, and then the followers determine their resource demands based on the selling prices, until reaching the utility equilibrium\cite{du2023beyond}. For example, the authors in \cite{zhong2023blockchain} focused on reliable vehicle twin migrations in vehicular metaverses and proposed a Stackelberg model between vehicular metaverse users and the roadside unit coalition with the highest utility.

\subsubsection{Contract theory}
Contract theory is a powerful tool for incentive mechanism design under information asymmetry, which has been effectively applied in IoT network optimizations\cite{du2023beyond}. Specifically, an employer, typically a service provider, designs contracts for specific tasks, and employees, i.e., the network users, engage in a contractual agreement\cite{du2023beyond}. For instance, the authors in \cite{wen2023freshness} studied Unmanned Aerial Vehicle (UAV)-enabled AI generative content and proposed a contract model. This model aimed to provide incentives for UAVs to contribute fresh data for GAI model fine-tuning under asymmetric information.

\subsubsection{Auction theory}
Auction theory focuses on studying the behavior and design of auctions for allocating resources through competitive bidding\cite{du2023beyond}. As an interdisciplinary technology, auction theory has been widely adopted for incentivizing resource trading in IoT networks, which can be implemented in asymmetric or incomplete information scenarios\cite{zhang2020auction}.  For example, the authors in \cite{zhang2020auction} proposed an auction-based optimization problem for the multichannel cooperative spectrum sharing in hybrid satellite-terrestrial IoT networks.

\subsection{Framework Design}
As shown in Fig. \ref{framework}, we introduce the GAI-based secure incentive mechanism framework for GIoT, which consists of a physical layer, an incentive layer, and a blockchain layer. We provide more details of the framework as follows: 


\begin{itemize}
    \item \emph{\textbf{Step 1. Design suitable incentive mechanisms:}} To address the reluctance of users to provide high-quality data for GAI model fine-tuning, edge servers as service providers would design suitable incentive mechanisms by considering the current conditions of GIoT networks, including network structures, performance metrics, resource constraints and so on.
    \item \emph{\textbf{Step 2. Adopt GDMs to derive optimal incentive mechanism design:}} Due to the ability to capture high-dimensional and complex structures of intricate environments, GDMs can be adopted to derive optimal incentive mechanism strategies that can maximize the utilities of edge servers\cite{du2023beyond}. The motivations and specific process of utilizing GDMs for designing efficient and robust incentive mechanisms are introduced in \cite{du2023beyond}.
    \item \emph{\textbf{Step 3. Issue the optimal incentive mechanism strategies:}} After finding the optimal incentive mechanism strategies, edge servers issue the strategies to the physical layer. Moreover, the resource trading involved in executing the strategies can be securely recorded and managed in the blockchain layer, ensuring transparency and security in resource trading.
    \item \emph{\textbf{Step 4. Obtain high-quality data for GAI model fine-tuning:}} Under the role of incentives, IoT devices collect fresh sensing data and provide them to the edge for GAI model fine-tuning. To ensure the quality of collected data, the reputation metric can be utilized to quantify the reliability of IoT devices and discourage malicious behavior, and the reputations of IoT devices would be securely managed in the blockchain layer\cite{zhong2023blockchain}.
    \item \emph{\textbf{Step 5. Enable GIoT applications:}} Based on the data collected by IoT devices or generated in the cloud, GAI model fine-tuning and inference can be performed on edge servers to efficiently enable GIoT applications, and the data stored on edge servers can be also securely managed in the blockchain layer. Considering that certain types of training data might be idle, GAI techniques have the ability to autonomously synthesize data, enhancing the performance of models\cite{ferrag2023generative}.
\end{itemize}

\begin{figure}[t]
\centerline{\includegraphics[width=0.45\textwidth]{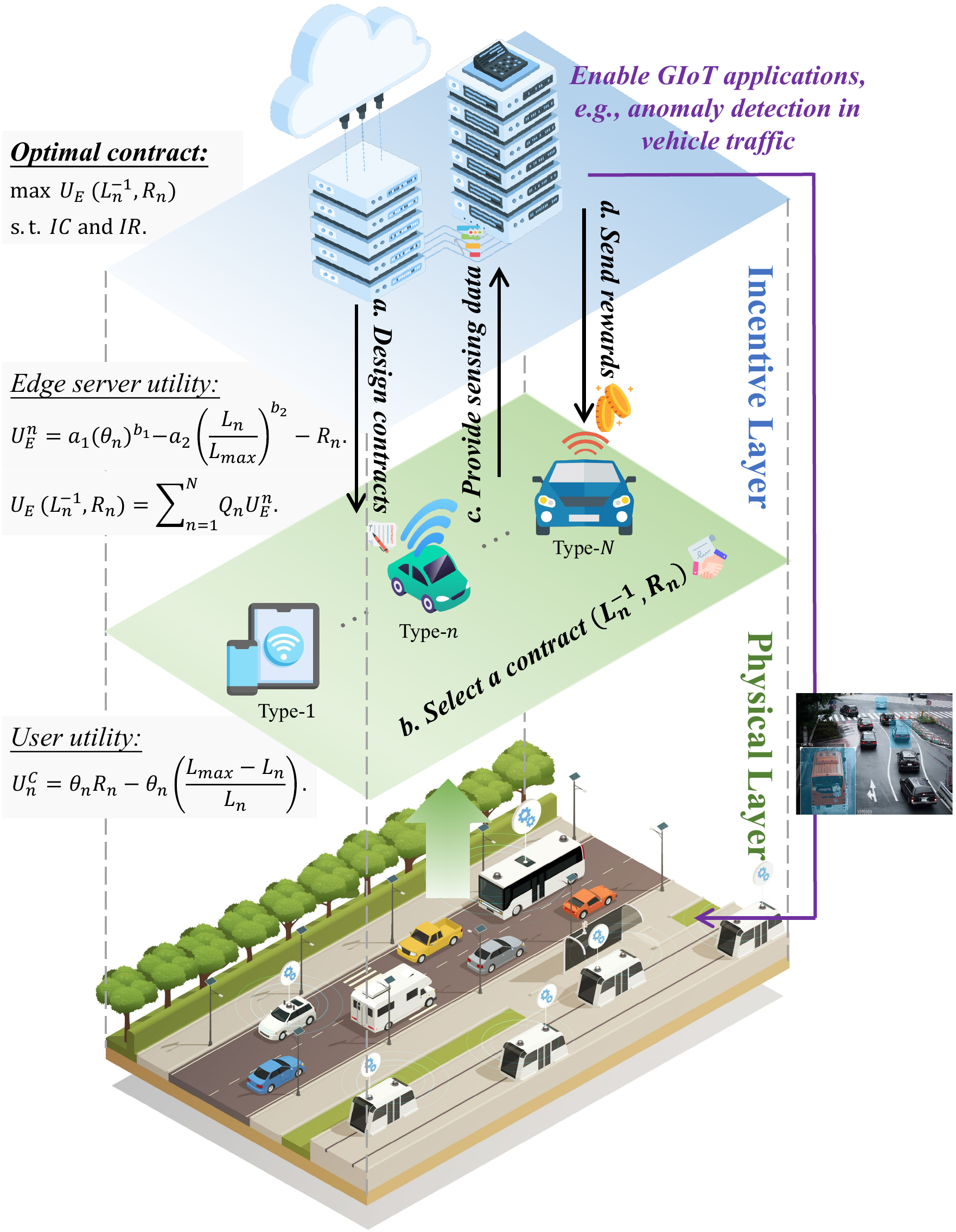}}
\captionsetup{font=footnotesize}
\caption{The illustration of GDM-based contract theory model for modern Internet of Vehicle traffic monitoring.}
\label{case}
\end{figure}

\section{Case Study: GDM-enabled Modern Internet of Vehicle Traffic Monitoring}
In this section, we present a case study on modern IoV traffic monitoring. Specifically, we propose a GDM-based contract theory model, which can incentivize users to contribute high-quality sensing data for GAI model fine-tuning, facilitating substantial advancements in intelligent transportation systems.

\subsection{System Model}
Figure \ref{case} depicts a specific case of the proposed framework. Specifically, with the capabilities of strong generalization and automated content generation\cite{xu2023unleashing}, GAI can offer personalized and advanced services to users, such as navigation and route optimization\cite{zhang2023generative}. To ensure the quality of services, GAI model fine-tuning at the edge requires high-quality datasets. However, due to the resource constraints of IoT devices, users may be reluctant to contribute fresh sensing data to the edge. Moreover, because of the dynamic and heterogeneous natures of vehicular networks\cite{zhang2023generative}, edge servers may lack awareness of the private information of users, such as their ability to collect sensing data, which can lead to users contributing data dishonestly to gain additional benefits\cite{wen2023freshness}. 

\subsection{Problem Formulation}
We consider that each edge server can support $M$ users. Based on statistical distributions of user types from historical data, we classify 
$M$ users into $N$ types and the user types are arranged in ascending order as $\theta_1 \leq \cdots \leq \theta_N$. In this definition, the higher type users can provide sensing data with the higher quality. For ease of understanding, the user with type $n$ is called the type-$n$ user.
\subsubsection{User utility}
The utility of type-$n$ users is denoted as $U_n^C$, equaling the difference between its obtained benefit and its cost of participation. As shown in Fig. \ref{case}, the obtained benefit of type-$n$ users is defined as $(\theta_nR_n)$\cite{kang2019toward}, where $R_n$ is the received reward. The cost of type-$n$ users is defined as $\theta_n(L_{max}/L_n-1)$\cite{liu2023deep}, where $L_n$ is the latency spent by type-$n$ users in collecting and transmitting sensing data with a guaranteed amount. Note that $L_{max}$ represents the highest permissible value of the latency.

\subsubsection{Edge server utility}
The utility obtained by the edge server from type-$n$ users is denoted as $U_E^n$, equaling the difference between the corresponding revenue for received datasets within $L_n$ and the reward $R_n$. According to \cite{liu2023deep,kang2019toward}, the revenue can be defined as a general quality-latency metric, i.e., $a_1(\theta_n)^{b_1}-a_2(L_n/L_{max})^{b_2}$. Here, $a_1>0$ and $a_2>0$ are pre-defined coefficients about the quality of received data and the latency spent for collecting and transmitting data\cite{kang2019toward}, respectively. Similarly, $b_1 \geq 1$ and $b_2 \geq 1$ are given factors measuring the effects of data quality and the latency\cite{kang2019toward}, respectively. Considering that the probability that a user is of type-$n$ is $Q_n$, where the sum of probabilities of all types is $1$, the expected utility of the edge server $U_E$ is shown in Fig. \ref{case}.


\subsubsection{Contract formulation}
As an economic tool, contract theory is effective in addressing information asymmetry for incentive mechanism designs\cite{wen2023freshness}. Therefore, the edge server can devise a contract comprising a group of contract items $(L_n^{-1}, R_n)$, where $L_n^{-1}$ is the reciprocal of $L_n$\cite{kang2019toward}. To ensure that each user optimally chooses the contract item designed for its type, the contract must satisfy both Individual Rationality (IR) and Incentive Compatibility (IC) constraints\cite{wen2023freshness}, where IR constraints indicate that the contract item that a user chooses should ensure a non-negative utility\cite{kang2019toward}, and IC constraints indicate that a user of any type prefers to choose the contract item designed for its type rather than any other contract item\cite{kang2019toward}. Finally, the optimization problem is to find the optimal contract $c^*$, i.e., $\{L_1^{-1^*},\ldots,L_N^{-1^*}\}$ and $\{R_1^{*},\ldots,R_N^{*}\}$, thereby maximizing the expected utility of the edge server $U_E$ while satisfying IR and IC constraints.

\subsection{GDM-empowered Contract Generation}
In this part, we adopt GDMs to derive optimal contract design\cite{du2023beyond}. Specifically, 
\begin{itemize}
    \item \textbf{Step 1. Model the environment state:} For simplicity, we consider that each edge server supports two types of users in the environment of vehicle traffic monitoring, where $\theta_1$ and $\theta_2$ are randomly sampled within $(10,100)$ and $(100,200)$, respectively\cite{liu2023deep}. Therefore, the environment state vector is defined as $S \triangleq [M, N, L_{max}, Q_1, Q_2, \theta_1, \theta_2]$. Note that $Q_1$ and $Q_2$ are generated randomly\cite{liu2023deep}, following the Dirichlet distribution, and $L_{max}$ is set as $150$.
    \item \textbf{Step 2. Formulate the participant utilities:} After determining the environment states, we formulate the utility of type-$n$ users $U_n^C$ and the expected utility of the edge server $U_E$, where the former is used to guarantee IR and IC constraints, and the latter is the optimization objective that we intend to maximize. Note that the weighting factors $a_1$, $a_2$, $b_1$, and $b_2$ are set as $15$, $10$, $1$, and $1$\cite{kang2019toward}, respectively.
    \item \textbf{Step 3. Customize the GDM settings:} The action space is the universe of the contract design\cite{du2023beyond}. Each contract is formed as $\{L_1^{-1}, R_1, L_2^{-1}, R_2\}$. Then, we can customize the model hyperparameters. For instance, in our case, the training epoch of the GDM is set as $120$, and the discount factor is set as $0.95$.
    \item \textbf{Step 4. Train the GDM and generate the optimal contract:} We train the policy $\pi(c^*|s)$ for generating the optimal contract $c^*$ under the state $s\in\{S\}$\cite{liu2023deep}. To evaluate each generated contract, we adopt the action-value function $Q(c^*|s)$\cite{liu2023deep}, which can guide the diffusion process. Finally, we can obtain the optimal contract $c^*$.
\end{itemize}

\begin{figure}[t]
\centerline{\includegraphics[width=0.45\textwidth]{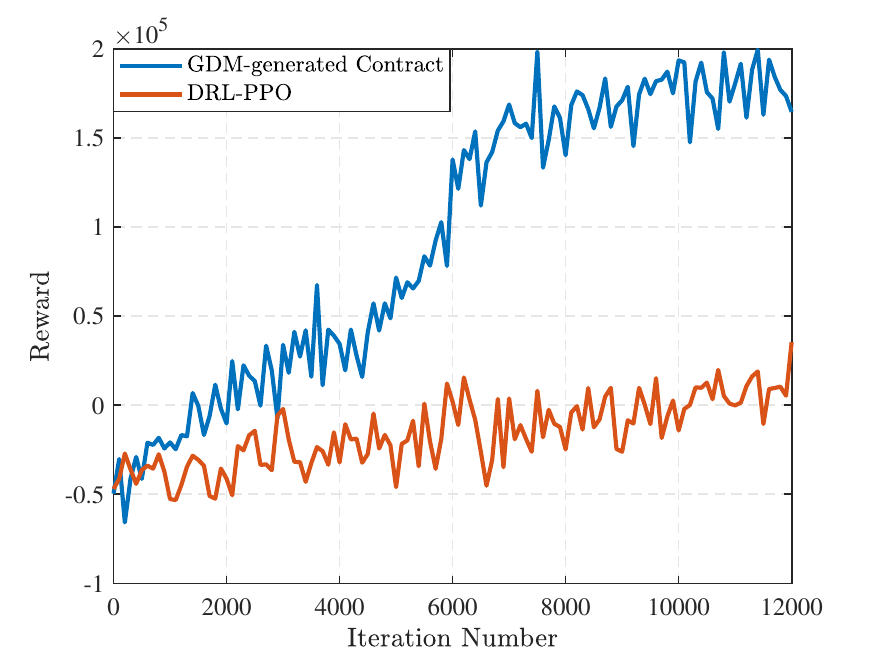}}
\captionsetup{font=footnotesize}
\caption{Training process of the GDM-based contract generation scheme and DRL-PPO for the optimal contract finding task, where the diffusion step is $100$, the batch size is $512$, and the contract generation network learning rate and contract quality network learning rate are $2\times10^{-7}$.}
\label{diffusion}
\end{figure}

\subsection{Numerical Results}
\begin{figure}[t]
\centerline{\includegraphics[width=0.45\textwidth]{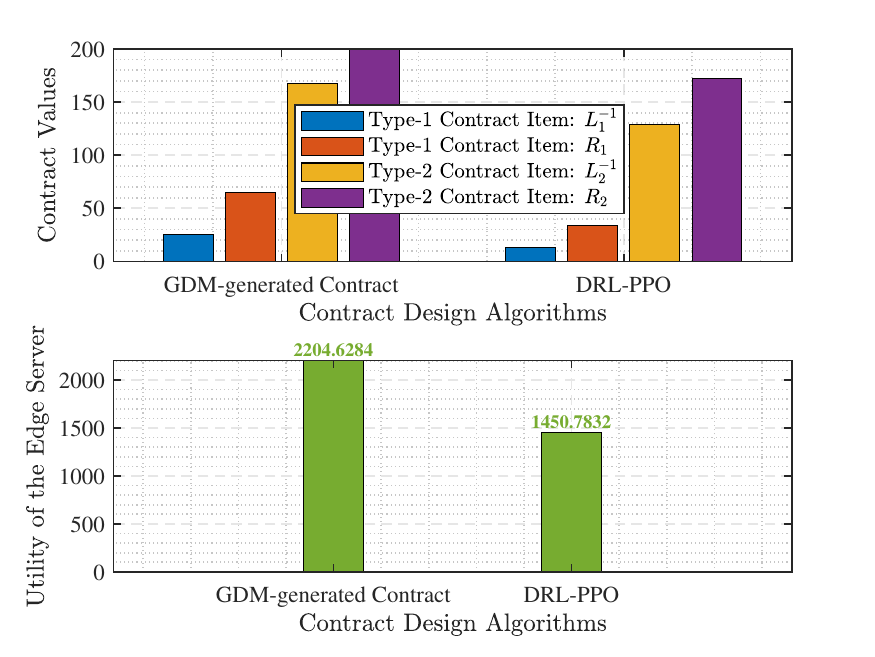}}
\captionsetup{font=footnotesize}
\caption{The designed contract comparisons between DRL-PPO and the GDM-based contract generation scheme.}
\label{contract}
\end{figure}

We carry out the experiment on NVIDIA GeForce RTX 3080 Laptop GPU with CUDA 12.0. Figure \ref{diffusion} shows test reward curves of our proposed GDM-based contract generation scheme and conventional DRL-PPO for the optimal contract finding task. We can observe that our proposed scheme always outperforms DRL-PPO under the same parameter settings. The reason is that the contract generation policy in our scheme is fine-tuned by the diffusion process, which can mitigate the impact of randomness and noise\cite{du2023beyond}. Figure \ref{contract} illustrates the quality of contracts generated by the proposed scheme and DRL-PPO. For a given environment state, we can observe that the proposed GDM-based contract generation scheme can provide a contract design that achieves the edge server utility value of $2204.6284$, which is greater than the $1450.7832$ achieved by the PPO. Overall, the above numerical results demonstrate that the performance of the proposed GDM-based contract generation scheme is better than that of DRL-PPO.

\section{Future Directions}
\subsection{Distributed and Green Generative AI Models}
One of the main challenges of GIoT for future development is the computational and storage limitations in training and deploying GAI models. For instance, GPT-3, as the state-of-the-art language model of OpenAI, consists of 175 billion parameters\cite{xu2023unleashing}, which is one of the largest language models in existence. Therefore, how to reduce the energy consumption of GAI models during their training and deployment is worth studying. One of the potential solutions is to design lightweight GAI models or adopt federated learning to train GAI models. 

\subsection{Quality Metrics for Reliable Generative AI Outputs}
Although GAI techniques have the incredible ability to automate content generation, they can be exploited to generate incorrect or fraudulent content, such as fake videos or wrong texts\cite{ferrag2023generative}. To address this issue, future research can explore the Quality of Service (QoS) metric from the user perspective to measure user satisfaction with the generated content. For instance, user preferences and feedback can be incorporated into the model training process. With the help of QoS, the performance of GAI models can be improved, and high-quality content can be generated to meet user satisfaction.

\subsection{Service Optimization by Prompt Engineering}
Formulating technical prompts to effectively instruct GIoT presents a challenge for individuals lacking adequate training in the relevant domain. Furthermore, the utilization of subpar prompts may diminish the overall generation quality of GAI models. Therefore, the exploration of prompt engineering for achieving the optimization of AI-generated content services is also a topic worthy of investigation. For instance, users can manually formulate diverse prompts and subsequently search for the one that yields the highest quality of generated outputs.

\subsection{Security and Privacy Protection for Users}
Centralized training or fine-tuning of GAI models at the edge may raise user concerns about data privacy and security\cite{xu2023unleashing}, as IoT data involving sensitive and personal information could potentially be exposed to attackers, leading to threats to user privacy and security. Therefore, future research can develop a user-centric privacy-preserving training approach to protect user security and privacy.

\section{Conclusion}
In this article, we presented the concept of \textit{Generative IoT (GIoT)}. Firstly, we reviewed several GAI techniques and explored their potential for IoT applications. Then, we summarized GIoT applications, including vision-based, audio-based, and text-based applications, and discussed the main challenges of integrating GAI with modern IoT to enable GIoT. To address these challenges, a general GAI-based secure incentive mechanism framework was proposed, in which we adopted GDMs for the optimal incentive mechanism design and utilized blockchain technologies for secure GIoT management. Furthermore, we conducted a case study on modern IoV traffic monitoring, leveraging GDMs to generate flexible contracts for motivating users to provide high-quality data for GAI model fine-tuning. The numerical results demonstrated the effectiveness of our proposed GDM-based contract generation scheme compared to DRL-PPO. Finally, we discussed potential research directions that can further facilitate the development of the GIoT ecosystem.

\bibliographystyle{IEEEtran}
\bibliography{ref}

\begin{thebibliography}{10}
\providecommand{\url}[1]{#1}
\csname url@samestyle\endcsname
\providecommand{\newblock}{\relax}
\providecommand{\bibinfo}[2]{#2}
\providecommand{\BIBentrySTDinterwordspacing}{\spaceskip=0pt\relax}
\providecommand{\BIBentryALTinterwordstretchfactor}{4}
\providecommand{\BIBentryALTinterwordspacing}{\spaceskip=\fontdimen2\font plus
\BIBentryALTinterwordstretchfactor\fontdimen3\font minus \fontdimen4\font\relax}
\providecommand{\BIBforeignlanguage}[2]{{%
\expandafter\ifx\csname l@#1\endcsname\relax
\typeout{** WARNING: IEEEtran.bst: No hyphenation pattern has been}%
\typeout{** loaded for the language `#1'. Using the pattern for}%
\typeout{** the default language instead.}%
\else
\language=\csname l@#1\endcsname
\fi
#2}}
\providecommand{\BIBdecl}{\relax}
\BIBdecl

\bibitem{du2023beyond}
H.~Du, R.~Zhang, Y.~Liu, J.~Wang, Y.~Lin, Z.~Li, D.~Niyato, J.~Kang, Z.~Xiong, S.~Cui \emph{et~al.}, ``Beyond deep reinforcement learning: A tutorial on generative diffusion models in network optimization,'' \emph{arXiv preprint arXiv:2308.05384}, 2023.

\bibitem{zhang2023complete}
C.~Zhang, C.~Zhang, S.~Zheng, Y.~Qiao, C.~Li, M.~Zhang, S.~K. Dam, C.~M. Thwal, Y.~L. Tun, L.~L. Huy \emph{et~al.}, ``A complete survey on generative {AI} ({AIGC}): Is {ChatGPT} from {GPT-4} to {GPT-5} all you need?'' \emph{arXiv preprint arXiv:2303.11717}, 2023.

\bibitem{9509294}
D.~C. Nguyen, M.~Ding, P.~N. Pathirana, A.~Seneviratne, J.~Li, D.~Niyato, O.~Dobre, and H.~V. Poor, ``{6G Internet of Things}: A comprehensive survey,'' \emph{IEEE Internet of Things Journal}, vol.~9, no.~1, pp. 359--383, 2022.

\bibitem{zhang2023generative}
R.~Zhang, K.~Xiong, H.~Du, D.~Niyato, J.~Kang, X.~Shen, and H.~V. Poor, ``Generative {AI}-enabled vehicular networks: Fundamentals, framework, and case study,'' \emph{arXiv preprint arXiv:2304.11098}, 2023.

\bibitem{ferrag2023generative}
M.~A. Ferrag, M.~Debbah, and M.~Al-Hawawreh, ``Generative {AI} for cyber threat-hunting in {6G}-enabled {IoT} networks,'' \emph{arXiv preprint arXiv:2303.11751}, 2023.

\bibitem{xia2023generative}
L.~Xia, Y.~Sun, C.~Liang, L.~Zhang, M.~A. Imran, and D.~Niyato, ``Generative {AI} for semantic communication: Architecture, challenges, and outlook,'' \emph{arXiv preprint arXiv:2308.15483}, 2023.

\bibitem{cao2023comprehensive}
Y.~Cao, S.~Li, Y.~Liu, Z.~Yan, Y.~Dai, P.~S. Yu, and L.~Sun, ``A comprehensive survey of {AI-Generated Content} ({AIGC}): A history of generative {AI} from {GAN} to {ChatGPT},'' \emph{arXiv preprint arXiv:2303.04226}, 2023.

\bibitem{liu2023deep}
Y.~Liu, H.~Du, D.~Niyato, J.~Kang, Z.~Xiong, D.~I. Kim, and A.~Jamalipour, ``Deep generative model and its applications in efficient wireless network management: A tutorial and case study,'' \emph{arXiv preprint arXiv:2303.17114}, 2023.

\bibitem{cook2019anomaly}
A.~A. Cook, G.~M{\i}s{\i}rl{\i}, and Z.~Fan, ``Anomaly detection for {IoT} time-series data: A survey,'' \emph{IEEE Internet of Things Journal}, vol.~7, no.~7, pp. 6481--6494, 2019.

\bibitem{xu2023unleashing}
M.~Xu, H.~Du, D.~Niyato, J.~Kang, Z.~Xiong, S.~Mao, Z.~Han, A.~Jamalipour, D.~I. Kim, V.~Leung \emph{et~al.}, ``Unleashing the power of edge-cloud generative {AI} in mobile networks: A survey of {AIGC} services,'' \emph{arXiv preprint arXiv:2303.16129}, 2023.

\bibitem{wen2023freshness}
J.~Wen, J.~Kang, M.~Xu, H.~Du, Z.~Xiong, Y.~Zhang, and D.~Niyato, ``Freshness-aware incentive mechanism for mobile {AI-Generated Content} ({AIGC}) networks,'' in \emph{2023 IEEE/CIC International Conference on Communications in China (ICCC)}.\hskip 1em plus 0.5em minus 0.4em\relax IEEE, 2023, pp. 1--6.

\bibitem{luong2016data}
N.~C. Luong, D.~T. Hoang, P.~Wang, D.~Niyato, D.~I. Kim, and Z.~Han, ``Data collection and wireless communication in {Internet} of {Things} ({IoT}) using economic analysis and pricing models: A survey,'' \emph{IEEE Communications Surveys \& Tutorials}, vol.~18, no.~4, pp. 2546--2590, 2016.

\bibitem{zhong2023blockchain}
Y.~Zhong, J.~Wen, J.~Zhang, J.~Kang, Y.~Jiang, Y.~Zhang, Y.~Cheng, and Y.~Tong, ``Blockchain-assisted twin migration for vehicular metaverses: A game theory approach,'' \emph{Transactions on Emerging Telecommunications Technologies}, p. e4856, 2023.

\bibitem{zhang2020auction}
X.~Zhang, D.~Guo, K.~An, G.~Zheng, S.~Chatzinotas, and B.~Zhang, ``Auction-based multichannel cooperative spectrum sharing in hybrid satellite-terrestrial {IoT} networks,'' \emph{IEEE Internet of Things Journal}, vol.~8, no.~8, pp. 7009--7023, 2020.

\bibitem{kang2019toward}
J.~Kang, Z.~Xiong, D.~Niyato, D.~Ye, D.~I. Kim, and J.~Zhao, ``Toward secure blockchain-enabled {Internet of Vehicles}: Optimizing consensus management using reputation and contract theory,'' \emph{IEEE Transactions on Vehicular Technology}, vol.~68, no.~3, pp. 2906--2920, 2019.

\end{thebibliography}

\end{document}